\documentclass{article}

\usepackage{arxiv}

\usepackage[utf8]{inputenc} % allow utf-8 input
\usepackage[T1]{fontenc}    % use 8-bit T1 fonts
\usepackage{hyperref}       % hyperlinks
\usepackage{url}            % simple URL typesetting
\usepackage{booktabs}       % professional-quality tables
\usepackage{amsfonts}       % blackboard math symbols
\usepackage{nicefrac}       % compact symbols for 1/2, etc.
\usepackage{microtype}      % microtypography
\usepackage{lipsum}
\usepackage{graphicx}
\graphicspath{ {./images/} }

\usepackage{natbib}
\usepackage{caption} % DO NOT CHANGE THIS AND DO NOT ADD ANY OPTIONS TO IT

\usepackage{algorithm}
\usepackage{algorithmic}

\usepackage[acronym,nonumberlist,section=chapter,toc, nogroupskip]{glossaries}
\newacronym{CB}{CB}{Contextual Bandits}
\newacronym{HMT}{HMT}{Human-Machine Team}
\newacronym{AI}{AI}{Artificial Intelligence}
\newacronym{RL}{RL}{Reinforcement Learning}
\newacronym{LinUCB}{LinUCB}{Linear Upper Confidence Bound}
\newacronym{UCB}{UCB}{Upper Confidence Bound}
\newacronym{DT}{DT}{Decision Tree}
\newacronym{ANN}{ANN}{Artificial Neural Network}
\newacronym{relu}{ReLU}{Rectified Linear Unit}
\newacronym{MSE}{MSE}{Mean Squared Error}
\newacronym{xAI}{xAI}{Explainable AI}
\newacronym{HMM}{HMM}{Hidden Markov Model}
\newacronym{ML}{ML}{Machine Learning}

\usepackage{amsfonts}
\usepackage{mathtools}
\usepackage{booktabs}

\title{Dynamic Trust Calibration Using Contextual Bandits}

\author{
 Bruno M. Henrique \\
  Thayer School of Engineering\\
  Dartmouth College\\
  Hanover, NH \\
  \texttt{bruno.miranda.henrique.th@dartmouth.edu} \\
   \And
 Eugene Santos Jr. \\
  Thayer School of Engineering\\
  Dartmouth College\\
  Hanover, NH \\
  \texttt{eugene.santos.jr@dartmouth.edu} \\
}

\begin{document}
\maketitle
\begin{abstract}
Trust calibration between humans and Artificial Intelligence (AI) is crucial for optimal decision-making in collaborative settings. Excessive trust can lead users to accept AI-generated outputs without question, overlooking critical flaws, while insufficient trust may result in disregarding valuable insights from AI systems, hindering performance. Despite its importance, there is currently no definitive and objective method for measuring trust calibration between humans and AI. Current approaches lack standardization and consistent metrics that can be broadly applied across various contexts, and they don't distinguish between the formation of opinions and subsequent human decisions. In this work, we propose a novel and objective method for dynamic trust calibration, introducing a standardized trust calibration measure and an indicator. By utilizing Contextual Bandits—an adaptive algorithm that incorporates context into decision-making—our indicator dynamically assesses when to trust AI contributions based on learned contextual information. We evaluate this indicator across three diverse datasets, demonstrating that effective trust calibration results in significant improvements in decision-making performance, as evidenced by 10 to 38\% increase in reward metrics. These findings not only enhance theoretical understanding but also provide practical guidance for developing more trustworthy AI systems supporting decisions in critical domains, for example, disease diagnoses and criminal justice.
\end{abstract}

% keywords can be removed
%\keywords{First keyword \and Second keyword \and More}

\section{Introduction}
\label{sec:intro}

Trust calibration can be defined as the process of achieving appropriate expectations regarding the capabilities of agents, whether humans or \gls{AI} autonomous systems \citep{trust-visser2014, trust-glikson2020, trust-naiseh2023, trust-ma2023}. With this definition in mind, incorrect levels of trust can lead to overreliance on flawed agents or, conversely, result in the dismissal of an agent with superior capabilities \citep{trust-dzindolet2003}. Hence, in the context of human-AI interaction for decision-making, trust calibration is a necessary condition for the most appropriate decision with respect to a given goal. 

Previous research has recognized the importance of trust calibration between humans and \gls{AI}. For instance, \citet{trust-visser2019} state that understanding the trust calibration process will lead to refinements in task division and coordination between team members. Also, there may be situations in which decisions should be handled entirely by humans if the \gls{AI} lacks sufficient accuracy to be trusted. Hence, trust calibration can help identify decisions that should be delegated to \gls{AI} or be scrutinized only by the human generalization capacity. Nevertheless, correct trust calibration is a challenge, particularly for individuals with limited domain knowledge, who have the burden of gauging the trustworthiness of the \gls{AI} system while estimating their own likelihood of being correct \citep{trust-ma2023}. 

An objective method to dynamically calibrate trust between humans and \gls{AI} would provide many advantages to a \gls{HMT}. For example, it would  alleviate the cognitive burden of assessing unknown \gls{AI} properties, such as the accuracy on a specific situation. Moreover, it would establish guardrails against potentially harmful agentic systems and still indicate situations when to adopt their opinions. Similarly, trust calibration works in the other way too, for example, indicating situations in which the human should make decisions with no \gls{AI} opinion. 

Despite the necessity of trust calibration for optimal \gls{HMT} performance, there is still no definite method for achieving it. Previous attempts include displaying uncertainty information alongside \gls{AI} recommendations \citep{trust-visser2014}, visual explanations \citep{trust-karran2022}, soliciting human input \citep{trust-gomez2023} and outcome feedback \citep{trust-ahn2024}. However, as concluded by \citet{trust-henrique2024}, \gls{AI} literature still lacks models and measures of trust calibration, especially algorithms and general implementation frameworks. In this context, this paper presents a dynamic trust calibration framework, including a calibration measurement definition and a dynamic indicator leading to optimal calibration. The framework is designed to be agnostic of the internal workings of \gls{AI} technology, broadening its applicability to any decision-support system.

Our proposal for a dynamic trust calibration indicator is grounded on multi-armed \gls{CB} algorithms. Like vanilla multi-armed bandits, \gls{CB} also present multiple arms/choices, but at each trial the context information is relevant for the decision. In this case, the agent making a decision does not change the state, as it happens in a typical \gls{RL} scenario, but the context changes. As such, \gls{CB} algorithms fit perfectly the framework of an external trust calibration process (the bandit), with the arms representing the indications to trust or distrust agents' opinions based on context variables, but with no changes in state. Despite the diverse applications of \gls{CB} \citep{mdp-li2010}, we are unaware of any previous research applying \gls{CB} to implement dynamic trust calibration.

% It is worth emphasizing that our proposal is a conceptual advance, providing an objective, performance-based measure of trust calibration that can be computed post-hoc based on decisions and their outcomes, not relying on subjective trust surveys or behavioral proxies alone. With that, it is not our goal to uncover or test human cognitive or affective trust states, but to quantify the performance improvement with calibrated trust. Therefore, we defer to future studies to determine how to utilize and communicate the proposed indicator.

It is worth emphasizing that our proposal is a conceptual advance, providing an objective, performance-based measure of trust calibration that can be computed post-hoc based on decisions and their outcomes. With that, it is not our goal to uncover or test human cognitive or affective trust states, but to quantify the performance improvement with calibrated trust. However, we acknowledge that embedding the indicator into real-time decision-support tools and evaluating its impact on actual user behavior and trust remains an important future research direction.

We evaluate our trust calibration indicator using empirical data obtained from recent studies in decision-making with \gls{AI} assistance. The data cover diverse domains and include multiple treatments, demonstrating the robustness of the proposed framework. Specifically, we compare the rewards accumulated by the actions informed by the trust calibration indicator with the total rewards obtained in the original studies. In all instances, the dynamic trust calibration indicator yields higher rewards, indicating improved trust calibration that enhances performance.

\section{Related work}
\label{sec:review}

Trust in \gls{AI} research originates from foundational studies on human trust in automation systems \citep{trust-henrique2024}. Many authors conducted experiments on human reliance on automation \citep{trust-dzindolet2003, trust-visser2016} and various theoretical trust models were proposed \citep{trust-lee2004, trust-visser2014, trust-hoff2014}. Other papers sought the factors that influence human trust in autonomous systems \citep{trust-hancock2011, trust-culley2013}. More recently, literature has shifted attention to explicitly studying measures of trust between humans and \gls{AI} \citep{trust-naiseh2023, trust-li2023, trust-leichtmann2023, trust-gomez2023}. However, as concluded by \citet{trust-henrique2024}, the literature still lacks objective methods to calibrate trust between humans and AI.

Some attempts to model trust calibration between humans and \gls{AI} rely on subjective actions. For example, the theoretical model of \citet{trust-visser2019} proposes to calibrate trust with actions such as apologies, denials, future promises, and warnings aimed at repairing damaged trust or reduce overtrust. Similar actions are explored in experiments by \citet{trust-pareek2024}, who demonstrate their influence on trust dynamics. However, these actions are inherently subjective and not directly translated by \gls{AI} recommender systems, which often function as "black-boxes".% for decision-makers. %Consequently, recalibration strategies based on apologies and promises are impractical in situations where the \gls{AI} implementation is not accessible. 

Recent studies on human-AI trust report experiments measuring reliance as a proxy to trust \citep{trust-yu2017}. For instance, \citet{trust-lu2021} study human reliance on \gls{AI} under conditions of limited accuracy information, revealing that the level of agreement between humans and \gls{AI} is very influential on reliance. Also measuring reliance as proxy to trust, \citet{trust-zhang2020} show that displaying the \gls{AI}'s confidence at each task can improve trust calibration, but they did not register improvements in performance. Confidence is also examined by \citet{trust-ma2024}, who concluded that correct human confidence calibration is necessary for appropriate reliance on \gls{AI} and, consequently, improved performance. However, correct confidence calibration can be as formidable a challenge as trust calibration.

Reliance on \gls{AI} recommendations is often viewed as trust behavior in the literature. For instance, \citet{trust-li2023} model human trust levels as a \gls{HMM}, using the hidden states as inputs to a likelihood function that predicts the human trust behavior of relying on \gls{AI} recommendations. By their turn, \citet{trust-wang2022} model the human trust behavior based on a reliance utility function. Their experiments suggest that trust calibration is influenced by the stakes of the decision. Also experimenting with trust behavior with respect to \gls{AI} recommendations, \citet{teams-noti2023} achieve improved team performance by determining when \gls{AI} opinions should be presented or omitted to human decision-makers. However, this approach requires access to the recommender system to include an optimization process responsible to advise when to show \gls{AI} opinions. %Moreover, this additional process introduces complexity to trust calibration, since \gls{AI} opinions would not be always available during consensus building.

Regardless of the trust calibration approach proposed, the usual objective of empirical human-AI trust research is to increase the \gls{HMT} performance on a given task \citep{trust-ahn2024, trust-salimzadeh2024, trust-ma2024, trust-gomez2023, trust-ma2023}. As such, it is expected that, when teamed together, humans and \gls{AI} will perform better than acting each by themselves in isolation. To this end, trust is necessary for team members to rely on the expertise of each other, with the goal of achieving complementary performance.% However, that goal can only be achieved by calibrating expectations with respect to each member's capabilities, that is, they must learn when a suggested opinion can be trusted or not. 

As a meaningful advance to complementary performance, \citet{trust-madras2018} introduced learning to defer, a framework in which an automated model learns to adaptively defer decisions to an external decision maker (DM) by considering the DM’s variable accuracy and biases. This approach generalizes traditional rejection learning by optimizing system-wide accuracy and fairness within a two-stage decision pipeline, where the DM is treated as a fixed and uncontrollable agent. While this method effectively addresses fairness concerns in static decision contexts and models the deferral decision as part of the learned predictor, it assumes a largely fixed interaction pattern between the model and the DM. As described next, our \gls{CB} takes a fundamentally different approach by leveraging all individual agent opinions at every decision point to estimate an optimal opinion, instead of deferring to one party exclusively.

% As briefly reviewed, trust in human-AI collaboration can be approached from many angles. However, there is no method to reliably measure trust directly. Some experiments consider self-reported scores of trust \citep{trust-leichtmann2023}, while others consider reliance as trust behavior and a proxy to trust \citep{trust-ma2024, trust-ahn2024, trust-li2023}. However, reported trust does not always correlate with reliance, and acceptance of a system recommendation does not necessarily mean trust. Acknowledging this caveat, many authors consider both, reported trust and trust behavior, in their experiments \citep{trust-salimzadeh2024, trust-gomez2023, trust-naiseh2023, trust-ma2023, trust-yu2017}. In the present case, our trust calibration framework does not rely on a direct trust measure. As described in the following section, we define a trust calibration measure based on team performance metrics.

% For more in-depth reviews on trust in \gls{AI}, we recommend the works of \citet{trust-henrique2024} and \citet{trust-glikson2020}.

\section{Methods}
\label{sec:methods}

We propose a framework for trust calibration in the traditional model of human and \gls{AI} agents working as a team, referred to as \gls{HMT}. In this model, the team goal is to make the best possible decision at each time/trial $t = \{1, 2, ..., n\}$ with respect to the context features, represented by a vector of $j$ independent variables $\mathbf{x} = \{x_1, x_2, ..., x_j\}$ that change at each $t$ in a sequence of decisions $\tau$. Before each decision, the team builds a consensus (absolute or relative), after considering the opinions from every member. Consequently, the consensual team opinion $o$ will lead to a decision and the corresponding perception of a reward $r(o_t, \mathbf{x}_t)$ for the time/trial $t$. However, this perceived reward may not be the maximum $R(o^*_t, \mathbf{x}_t)$, only perceived by making the decision corresponding to an optimal opinion $o^*$. Therefore, $r(o_t, \mathbf{x}_t) \leq R(o^*_t, \mathbf{x}_t)$. In this scenario, there are two potentially distinct quantities: the total amount of rewards possible with optimal opinions for the entire sequence presented to the \gls{HMT} up until $t$, denoted by $G(\tau_t) = \sum_t R(o^*_t, \mathbf{x}_t)$, and the actual amount of rewards obtained by the \gls{HMT}, denoted by $g(\tau_t) = \sum_t r(o_t, \mathbf{x}_t)$, such that $g(\tau_t) \leq G(\tau_t)$ \footnote{In this paper, we utilize the notation "o" (opinion), contrasting with the traditional Reinforcement Learning literature, which commonly uses "a" (action). This distinction is intentional and essential, as our framework emphasizes the need to differentiate between the team’s actions and the trust calibration process, which relies on opinions formulated prior to those actions.}.

Following previous definitions, trust calibration is the process of adjusting the agents' levels of trust in other agents' opinions with respect to their capabilities \citep{trust-visser2014, trust-glikson2020, trust-naiseh2023, trust-ma2023}. As such, correctly calibrated trust levels mean anticipating when the opinions of the agents should be trusted or not. For instance, in a human-AI collaboration scenario of sequential decision-making in which the human makes each decision after considering the \gls{AI} opinion, the human calibrates their trust levels with respect to the perceived \gls{AI} capability for correct opinions. Specifically, as the team collaborates towards maximizing the number of optimal decisions, the human learns the contexts (i. e., the set of variables for each decision) in which the \gls{AI} recommender system should be trusted or not. In this example, perfectly calibrated trust means that the human can tell when the \gls{AI} is right or wrong, that is, when to trust it or not in order to achieve the optimal reward.

It should be noted that in the previous paragraph (and throughout the rest of this paper) we refer to trust as behavioral trust, that is, observed reliance on \gls{AI} opinions. Following \citet{trust-ma2024}, \citet{trust-ahn2024} and \citet{trust-li2023}, we use reliance as a proxy to trust since there is no definite measurement of the agents' inner models of trust with respect to their teammates. As such, we rely on the agents' behavior as an indicator of trust in their teammates' opinions: if an agent trusts an opinion, they will follow it; otherwise, they will modify their decision. In this context, trust calibration does not translate into just following others' opinions (for example, a human blindly following the \gls{AI} recommender system). It refers to adjusting the behavioral trust to accept opinions perceived as optimal and deny opinions perceived as wrong or sub-optimal. Also, trust calibration should not be confused with individual agents' performance (such as \gls{AI} accuracy). %In fact, it is the process of adapting trust behavior to the perceived performance of individual agents.

% With the considerations above, we propose a measure of trust calibration by noticing that $R(o^*, \mathbf{x})$ is only perceived when the \gls{HMT} correctly identifies which individual member's opinions should be trusted or not with respect to the decision variables $\mathbf{x}$, and appropriately follows $o^*$. That is, $R(o^*, \mathbf{x})$ is perceived when trust is calibrated, regardless of how the consensus was achieved (it might be just the case of a human making the final decision considering an \gls{AI} opinion). Conversely, if trust is not calibrated, one or more wrong opinions that should not be trusted will lead the \gls{HMT} to decide $o$ sub-optimally, perceiving $r(o, \mathbf{x})$. Therefore, trust miscalibration can be quantified by the regret of the team in choosing to follow one or more wrong opinions. More formally, at any point $t$ during the sequence of decisions $\tau$, we define the trust calibration distance $T(\tau_t)$ as the difference between the total perceived rewards with optimal team opinions $o^*$ and the actual rewards with team opinions $o$, up to time/trial $t$:

With the considerations above, we propose a measure of trust calibration by noticing that $R(o^*, \mathbf{x})$ is only perceived when the \gls{HMT} correctly identifies which individual member's opinions should be trusted or not with respect to the decision variables $\mathbf{x}$, and appropriately follows $o^*$. That is, $R(o^*, \mathbf{x})$ is perceived when trust is calibrated, regardless of how the consensus was achieved. Conversely, if trust is not calibrated, one or more wrong opinions that should not be trusted will lead the \gls{HMT} to decide $o$ sub-optimally, perceiving $r(o, \mathbf{x})$. Therefore, trust miscalibration can be quantified by the regret of the team in choosing to follow one or more wrong opinions. More formally, at any point $t$ during the sequence of decisions $\tau$, we define the trust calibration distance $T(\tau_t)$ as the difference between the total perceived rewards with optimal team opinions $o^*$ and the actual rewards with team opinions $o$, up to time/trial $t$:

\begin{equation}
    \label{eq:trustcalibration}
    \begin{split}
        T(\tau_t) & \vcentcolon = \left \vert G(\tau_t) - g(\tau_t) \right \vert 
        \\ & = \left \vert \sum_t R(o^*_t, \mathbf{x}_t) - \sum_t r(o_t, \mathbf{x}_t) \right \vert.
    \end{split}
\end{equation}

The trust calibration distance can be interpreted as the amount of rewards lost due to trust miscalibration.\footnote{In the \gls{RL} paradigm, the trust calibration distance is given by the cumulative regret up until $t$.} It represents how far the agents were from perfect trust calibration in the past sequence of decisions $\tau_t$. As such, $T(\tau_t) = 0$ means that the agents had perfect trust calibration in a given sequence $\tau_t$: in this trajectory of decisions, they chose to trust each other's opinions only when they would result in the highest rewards (or lower regrets) possible. On the other hand, $T(\tau_t)$ increases with every sub-optimal decision $o$. 

The definition presented in Equation \ref{eq:trustcalibration} emphasizes that $T$ is a function of a given past sequence of decisions $\tau$. Hence, it is only possible to measure trust calibration regarding past decisions. Moreover, $T$ is a dynamic measure, changing with each new trial/time unit $t$, congruent with previous models of trust in the literature \citep{trust-yu2017}. It is also noteworthy that the trust calibration distance is independent of individual agents' performance in the \gls{HMT}. This is consistent with the definition of trust calibration as a process that adjusts trust to individual agents' capabilities, arriving at the decision of which agents' opinions to follow in each case. %For instance, the \gls{HMT} eventually calibrates trust to individual agents with poor performance, effectively learning when they should not be trusted, improving overall team performance measured by $g(\tau_t)$. With that, even teams with low-performing agents might achieve high levels of trust calibration, which will be denoted by low values of $T$.

\begin{figure}
    \centering
    \captionsetup{justification=justified}
    \includegraphics[width=7.90cm]{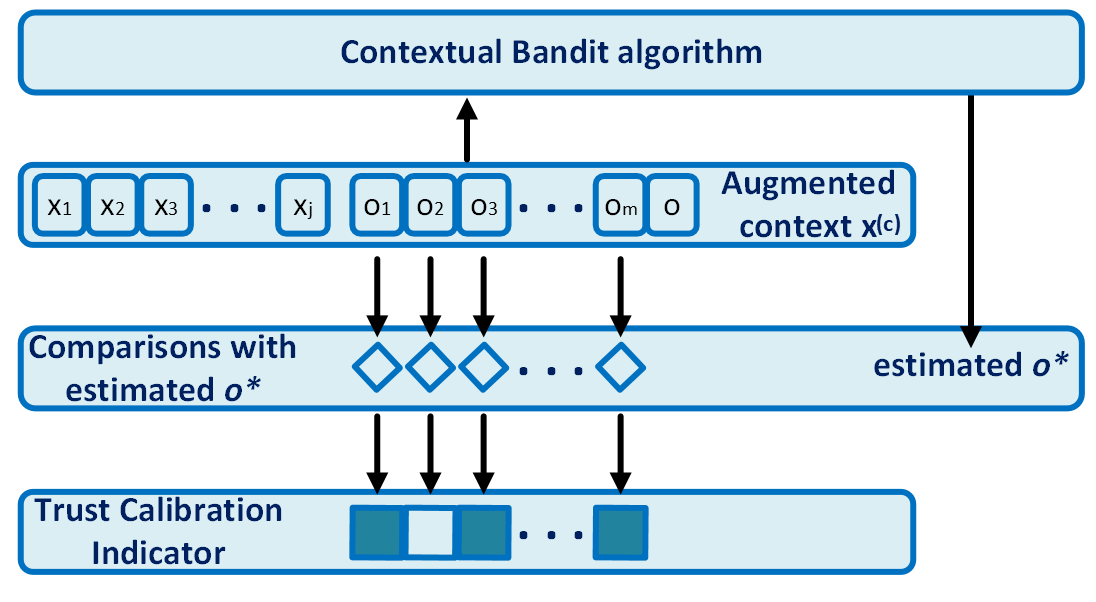}
    \caption{The augmented context $\mathbf{x}^{(c)}$ includes the original context variables $\mathbf{x} = \{x_1, x_2, x_3, ..., x_j \}$ and individual agent opinions depicted by $\{o_1, o_2, o_3, ..., o_m\}$, as well as the consensual team opinion $\{o\}$ before the decision. This augmented context is used by the \gls{CB} algorithm to estimate $o^*$, which is used to calibrate trust with regards to individual opinions. Opinions comparisons to the estimated $o^*$ are represented by diamonds, indicating to trust an individual opinion (clear squares) or not (dark squares).}
    \label{fig:indicatorframeworkdetail2}
\end{figure}

As Equation \ref{eq:trustcalibration} implies, the trust calibration distance $T$ is reduced when $g(\tau_t) \rightarrow G(\tau_t)$, meaning the \gls{HMT} has improved trust calibration on that sequence of decisions $\tau_t$. In this case, the team calibrated their decision to trust or distrust each others' opinions in each trial/time $t$, with respect to the decision variables $\mathbf{x}$, achieving $g(\tau_t)$ close (or equal) to $G(\tau_t)$. Specifically, by definition, the team calibrates trust toward the optimal opinion $o^*$ by deciding which individual agents opinions, denoted by $o_1, o_2, ..., o_m$, should be trusted or not in each $t=\{1, ..., n\}$ with respect to the respective decision variables $\mathbf{x}_t$. With that in mind, we build a trust calibration indicator by estimating the $o^*$ before every decision and evaluating individual agents' opinions relative to it (Figure \ref{fig:indicatorframeworkdetail2}). If those individual opinions diverge from the estimated $o^*$, they are deemed distrustful by the indicator at $t$. On the other hand, individual opinions that align with the estimated $o^*$ at $t$ will be indicated as trusted. Therefore, the proposed trust calibration indicator shows the trust behavior more likely to result in the optimal decision, leading to $g(\tau_t) \rightarrow G(\tau_t)$ as intended.

At this point, it should be noted that the trust calibration indicator is dynamic, changing for every opinion in $t$ and with respect to the decision variables $\mathbf{x}_t$ and the individual agents' opinions $o_1, o_2, ..., o_m$ in an \gls{HMT}. As such, individual opinions in a team are labeled to be trusted or not in the very specific circumstance of $t$, not for the whole sequence $\tau$. With this, the trust calibration indicator is agnostic of perceptions of trustworthiness of individual agents, that is, the agents themselves are not judged as trustful or distrustful whatsoever. The only evaluations made by the indicator are regarding individual opinions relative to the estimated $o^*$ at hand, regardless of individual characteristics of the agents (such as accuracy). Therefore, the indicator refers to both humans' and artificial agents' opinions alike. Also, the indicator is calculated before the \gls{HMT} decision action is issued and the reward is perceived.\footnote{We use $o_1, o_2, ..., o_m$ to distinguish the individual opinions of the $m$ agents in a team from the consensus team opinion given by $o$. Notice that, depending on the scenario, $o$ might be the final call made by a human, as in traditional decision-making assisted by \gls{AI}.}

From the trust calibration indicator perspective, $o^*$ is estimated according to a dynamically changing context, before the team decision is issued and the outcome is known. Therefore, the trust calibration indicator considers the context features variables presented to every individual agent of the \gls{HMT}, as well as their respective opinions, and treats the whole as a set of input features necessary to estimate $o^*$. It should be noted that the context features presented to the \gls{HMT}, as well as the team and individual opinions, are all considered as a whole set of inputs for trust calibration. This set of inputs is called augmented context, defined as

\begin{equation}
    \label{eq:augmentedcontext}
    \mathbf{x}^{(c)} \vcentcolon = \{x_1, x_2, ..., x_j, o_1, o_2, ..., o_m, o\}.
\end{equation}

The trust calibration indicator relies on estimating $o^*$ before the \gls{HMT} action. This estimative is given by a model capable of dynamically mapping situations (context) to opinions in order to maximize rewards, such as the \gls{RL} paradigm introduced by \citet{mdp-sutton1998}. In traditional \gls{RL}, an agent in a given state $s$ selects an action $a$, transitions to a new state $s'$ and receives a reward $r(s, a, s')$. However, in the present case of the trust calibration indicator, each action does not involve state transitions. Hence, our framework fits the \gls{RL} sub-field of multi-armed \gls{CB}, in which the estimation of $o^*$ is only based on (augmented) contextual information $\mathbf{x}^{(c)}$ and the corresponding reward \citep{mdp-lu2010}.

A classic stochastic multi-armed bandit problem is defined as $K$ possible actions of a player, each associated with a probability distribution of rewards $D_1, D_2, ..., D_K$, with respective means and variances unknown to the player. The usual problem description uses a slot machine analogy, with $K$ arms available to the player, who has to pull one at each time/trial $t$ \citep{mdp-kuleshov2014}. Since the parameters of the rewards distributions $D_1, D_2, ..., D_K$ are unknown, multi-armed bandit algorithms are used to uncover strategies for players to maximize their rewards (or minimize their regret). However, when relevant context information is available to the player, the problem becomes multi-armed \gls{CB}, as defined by \citet{mdp-langford2007}. In the present case, we use the multi-armed \gls{CB} model to estimate $o^*$ among the $K$ possible opinions of actions, using the augmented context $\mathbf{x}^{(c)}$.

There are many \gls{CB} algorithms to estimate an optimal opinion $o_t^*$ considering the context information $\mathbf{x}_t^{(c)}$ \citep{mdp-silva2022, mdp-kuleshov2014}. After this estimation, it is just a matter of comparing individual opinions $o_1, ..., o_m$ to the estimated $o^*$ to have the indications of which agents should be trusted in this particular circumstance $t$, completing the trust calibration indicator (Figure \ref{fig:indicatorframeworkdetail2}). For the purposes of demonstrating our approach to dynamic trust calibration, we select three popular algorithms to estimate $o^*$: \gls{LinUCB} \citep{mdp-li2010}, \gls{DT} bandits \citep{mdp-elmachtoub2017} and \gls{ANN} bandits \citep{mdp-kakade2008}. %They are briefly reviewed in the following sub-section.

\subsection{Contextual Bandits algorithms}
\label{sec:methods:cb}

The \gls{LinUCB} approach to \gls{CB} was originally proposed by \citet{mdp-li2010}. It assumes that the expected payoff of an opinion $o$ is linear with respect to the augmented context features $\mathbf{x}^{(c)}$ and unknown coefficients $\boldsymbol{{\theta}}_o$ for each opinion/arm $o \in \{o_1, ..., o_K\}$, that is,

% \begin{equation}
%     \label{eq:linucb1}
%     \mathbb{E} \left[ r_{t,o} | \mathbf{x}^{(c)}_{t,o} \right] = \left( \mathbf{x}^{(c)}_{t,o} \right)^\top \boldsymbol{\theta}_o.
% \end{equation}

\begin{equation}
    \label{eq:linucb1}
    \mathbb{E} \left[ r_{t,o} | \mathbf{x}^{(c)}_{t} \right] = \left( \mathbf{x}^{(c)}_{t} \right)^\top \boldsymbol{\theta}_o.
\end{equation}

In order to estimate $o^*$ at each $t$, an \gls{UCB} is calculated for each possible opinion/arm $o$ as 

\begin{equation}
    \label{eq:linucb2}
    \mbox{UCB}_{o, t} \left( \mathbf{x}^{(c)}_t \right) = \boldsymbol{\theta}_o^\top \mathbf{x}^{(c)}_t + \sqrt{ \left( \mathbf{x}^{(c)}_t \right)^\top \mathbf{A}_o^{-1} \mathbf{x}^{(c)}_t},
\end{equation}

\noindent where $\mathbf{A}_o$ is a positive definite matrix defined as

\begin{equation}
    \label{eq:linucb3}
    \mathbf{A}_o \vcentcolon = \sum_{t:o_t = o}\mathbf{x}^{(c)}_t \left( \mathbf{x}^{(c)}_t \right)^\top.
\end{equation}

\noindent Then, the estimated $o^*$ at $t$ for a given $\mathbf{x}_t$ is selected as the $o$ that maximizes the \gls{UCB} given by Equation \ref{eq:linucb2}, that is,

\begin{equation}
    \label{eq:linucb4}
    \mbox{estimated } o_t^* = \arg\max_{o \in \{o_1, ..., o_K\}} \mbox{UCB}_{o,t} \left(\mathbf{x}^{(c)}_t \right).
\end{equation}

\noindent Assuming the algorithm has access to the reward $r_{t, o^*}$ from the estimated $o^*$, the matrix $\mathbf{A}_o$, in Equation \ref{eq:linucb3}, is updated as $\mathbf{A}_o = \mathbf{A}_o + \mathbf{x}^{(c)}_t \left( \mathbf{x}^{(c)}_t \right)^\top$. Similarly, the parameters $\boldsymbol{\theta}_o$, from Equation \ref{eq:linucb1}, are updated as $\boldsymbol{\theta}_o = \mathbf{A}_o^{-1} \left( \mathbf{x}^{(c)}_t \right)^\top r_{t,o^*}$.

\gls{DT}s were introduced in the multi-armed \gls{CB} problem by \citet{mdp-elmachtoub2017} because they are simple, interpretable and non-parametric, capable of capturing non-linear relationships between variables. The algorithm consists in using the original \gls{DT} proposed by \citet{finance-breiman1984} to estimate the $o^*$ given a vector of augmented context features variables $\mathbf{x}^{(c)}$. It splits the available previous data into a tree-like structure where each node represents a feature variable, the branches represent the decision rules with respect to the values of those variables, and the leaf nodes are the opinions. Hence, to decide which opinion is the estimated $o^*$ for the context variables vector $\mathbf{x}^{(c)}_t$, the algorithm parses $\mathbf{x}^{(c)}_t$ down the tree, deciding which branch to follow based on the values of each variable in $\mathbf{x}^{(c)}_t$, until the leaf containing the estimated $o^*$ is achieved. After perceiving $r_{t, o^*}$, the tree-like structure is updated to reflect this new data point.

\gls{ANN}s are universal approximators of arbitrary functions, particularly useful for complex and high-dimensional data. As such, they can be used as a \gls{CB} approach to approximate the relationship between context variables $\mathbf{x}^{(c)}$ and the estimated $o^*$, as demonstrated by \citet{mdp-kakade2008}. A vanilla neural network consists of fully interconnected neurons, organized in input and hidden layers, as well as output neurons. The neurons of the input layer correspond to the variables of vector $\mathbf{x}^{(c)}$, each connecting to one or more neurons in the hidden layer. The hidden layer neurons apply an activation function to their inputs, such as sigmoid, hyperbolic tangent or \gls{relu}, defined as $f(x) \vcentcolon = \max (0, x)$. Finally, the output layer neurons produce the estimated $o^*$ as a function of the input $\mathbf{x}^{(c)}$ at each $t$.

Many choices of architecture are possible for \gls{ANN}s, including multiple hidden layers for deep learning \citep{finance-schmidhuber2015}. To demonstrate the estimations of $o^*$ for the trust calibration indicator, we select a vanilla neural network \gls{CB} composed of an input layer fully connected with weights $\mathbf{w}$ to a single hidden layer with 15 neurons implementing \gls{relu}. Each hidden neuron $z_h$ receives the weighted sum of the input layer variables, such that $z_h = \sum_{i}w_{i,h}x_i$, for $h = \{1, ..., 15\}$ and all $i$ variables from vector $\mathbf{x}^{(c)}$ in Equation \ref{eq:augmentedcontext}. Since we select \gls{relu} as activation, the output layer is given by the linear combination of $f(z_h) = \max(0, z_h)$. Finally, for updating the \gls{ANN} with each new datapoint, we select the \gls{MSE} loss function and Adam optimization \citep{finance-kingma2015}.

\subsection{Data}
\label{sec:methods:data}

% We evaluate the trust calibration indicator over data from previous studies in decision-making assisted by \gls{AI}. Specifically, we use data from experiments conducted by \citet{trust-lu2021}, \citet{teams-noti2023} and \citet{trust-reverberi2022}. Those datasets were selected because of availability and diversity of domains and treatments, each with its own set of features. Hence, they show the broad applicability of our trust calibration framework. We consider the decisions of the original data as opinions of an \gls{HMT} where humans had the final call, but they were exposed to the \gls{AI} opinions. In this regard, even if the human decided to ignore the \gls{AI}, because they saw its opinion before making the decision, it is reasonable to assume that, in some cognitive level, the human took that opinion into consideration. 

We evaluate the trust calibration indicator over data from previous studies in decision-making assisted by \gls{AI}. Specifically, we use data from experiments conducted by \citet{trust-lu2021}, \citet{teams-noti2023} and \citet{trust-reverberi2022}. These datasets were selected due to their availability and diversity in domains, experimental treatments, and feature sets. In all cases, the teams consist of one human and one \gls{AI} agent, with the human having the final decision-making authority after considering the \gls{AI}’s opinion. We treat the recorded decisions in these studies as the individual opinions of this \gls{HMT} for our analysis. %In this regard, even if the human decided to ignore the \gls{AI}, because they saw its opinion before making the decision, it is reasonable to assume that, in some cognitive level, the human took that opinion into consideration. 

\citet{trust-lu2021} experiment with speed dating data, asking a total of 1,169 human subjects to predict at each trial $t$ if a person, according to a set of 22 known features $\mathbf{x} = \{x_1, x_2, ..., x_{22}\}$, would want to see their date again or not.\footnote{For illustration, features include attractiveness, intelligence, shared interests, sincerity, ambition. Please refer to \citet{trust-lu2021} for the full set of features.} After the human subjects have an initial binary opinion $o_1$, they are given an opinion from a \gls{ML} algorithm, which we refer to as the \gls{AI} opinion $o_2$. Finally, the \gls{HMT} achieves a consensual opinion $o$. Also, the authors evaluate different levels of agreement between humans and \gls{AI} in one phase and use that information to vary treatments in the next phase of experiments. Similarly, they also vary levels of human confidence in their own responses. % Within those factorial designs, in experiment 1 they do not show the \gls{AI} accuracy, whereas in experiment 2, humans have the accuracy knowledge. Finally, the third experiment reported by \citet{trust-lu2021} varies both human confidence in their decision when they agree with the \gls{AI} and their confidence when they disagree with the \gls{AI}. 

By their turn, \citet{teams-noti2023} work with the risk of released defendants violating the release terms. In their experiments, 1,096 human subjects are asked to estimate that risk, from 0 to 100\% in increments of 10\%, for $n$ released defendants. Similarly to other experiments, humans are aided by a risk assessment tool developed previously, which, for our purposes, will be considered the \gls{AI} recommender system issuing opinions $o_2$. Subjects observe a set of features $\mathbf{x} = \{x_1, x_2, ..., x_7\}$, then the \gls{AI} opinion $o_2$ is shown according to different treatments.\footnote{Features are: ethninicy, gender, age, offense type, number of prior arrests, number of previous failures to appear in court and number of prior convictions.} The treatments are: update, which shows the recommended $o_2$ after the humans had a first opinion $o_1$, giving them the opportunity to update their team opinion $o$; learned (the authors' novel approach), which shows the opinion $o_2$ only when their \gls{AI} algorithm estimates that the information might improve the final opinion $o$ (according to a previously trained policy); omniscient, very similar to the learned treatment, but the recommender system will always show the optimal suggestion to the human when it might improve the opinion $o$. We note that traditional human decision-making assisted by \gls{AI} fits the update treatment, but, for the sake of completion, we evaluate the trust calibration indicator over all treatments.\footnote{Learned and omniscient treatments impose conditions on the \gls{AI} recommender system that might not be realistic when the system is given as a black-box, with no access to its inner models.}

% Still with respect to the data from \citet{teams-noti2023}, it should be noted that the opinions in this case are not binary, but multi-valued. Specifically, the risk value given by $o_1$, $o_2$ and $o$ can be any integer $V \in \{0, 10, 20, ..., 100\}$. However, the trust calibration indicator is designed to show which individual agents should be trusted or not by comparing $o_1$, $o_2$ and $o$ to the estimated $o^*$. As such, just indicating trust behavior with respect to $o$, $o_1$ and $o_2$ might not be enough to decrease the trust calibration distance $T$ (Equation \ref{eq:trustcalibration}). For example, in \citet{teams-noti2023} dataset, the trust calibration indicator might deem $o_t$, $o_{1,t}$ and $o_{2,t}$ untrustworthy, which is not enough to reduce the trust calibration distance $T$ as intended. However, in cases like this, with non-binary opinions, the indicator can readily provide the estimated $o_t^*$ if no agent's individual opinion is to be trusted. Hence, instead of just highlighting which agents' opinions should be trusted, the indicator can also provide a better alternative opinion. This underscores an advantage of building the indicator from the estimated $o^*$, since this is the opinion that will mostly likely lead to the real $R(o_t^*, \mathbf{x}_t)$.

Still with respect to the data from \citet{teams-noti2023}, it should be noted that the opinions in this case are not binary, but multi-valued. Specifically, the risk value given by $o_1$, $o_2$ and $o$ can be any integer $V \in \{0, 10, 20, ..., 100\}$. However, the trust calibration indicator is designed to show which individual agents should be trusted or not by comparing $o_1$, $o_2$ and $o$ to the estimated $o^*$. As such, just indicating trust behavior with respect to $o$, $o_1$ and $o_2$ might not be enough to decrease the trust calibration distance $T$ (Eq. \ref{eq:trustcalibration}). For example, the trust calibration indicator might deem $o_t$, $o_{1,t}$ and $o_{2,t}$ untrustworthy, leaving no opinion to be trusted. However, in cases like this, with non-binary opinions, the indicator can readily provide the estimated $o_t^*$ if no agent's individual opinion is to be trusted. %Hence, instead of just highlighting which agents' opinions should be trusted, the indicator can also provide a better alternative opinion. This underscores an advantage of building the indicator from the estimated $o^*$, since this is the opinion that will mostly likely lead to the real $R(o_t^*, \mathbf{x}_t)$.

% The final dataset we consider in this paper comes from the experiments of \citet{trust-reverberi2022}. They experiment with 21 endoscopists who are asked to diagnose colorectal lesions based on 504 real colonoscopies videos, each with one lesion (the diagnosis is either true or false). They first diagnose $o_1$ with no decision support system and, then, they receive a report from an \gls{AI} with a recommended diagnosis $o_2$. After considering the \gls{AI} recommendation, the endoscopists issue a final diagnosis $o$. Although the underlying \gls{AI} system had full access to the features that enable it to predict diagnoses, the dataset provided by \citet{trust-reverberi2022} does not list those features (the full set of $\mathbf{x}$ is unknown to our trust calibration indicator). The only variables $\mathbf{x} = \{x_1, x_2, ..., x_{11}\}$ available to the trust calibration indicator at each trial $t$ are relative to \gls{AI} and human confidence levels, endoscopists' expertise and their beliefs regarding the diagnoses.\footnote{Please refer to \citet{trust-reverberi2022} for the full set of features.} For the \gls{HMT}'s opinion $o$, we consider the endoscopists' diagnoses after they have seen the \gls{AI} opinions $o_2$.

The final dataset comes from the experiments of \citet{trust-reverberi2022}. They experiment with 21 endoscopists who are asked to diagnose colorectal lesions based on 504 real colonoscopies videos, each with one lesion (the diagnosis is either true or false). They first diagnose $o_1$ with no decision support system and, then, they receive a report from an \gls{AI} with a recommended diagnosis $o_2$. After considering the \gls{AI} recommendation, the endoscopists issue a final diagnosis $o$. The variables $\mathbf{x} = \{x_1, x_2, ..., x_{11}\}$ available to the trust calibration indicator at each trial $t$ are relative to \gls{AI} and human confidence levels, endoscopists' expertise and their beliefs regarding the diagnoses.\footnote{Please refer to \citet{trust-reverberi2022} for the full set of features.} For the \gls{HMT}'s opinion $o$, we consider the endoscopists' diagnoses after they have seen the \gls{AI} opinions $o_2$.

The selected datasets come from diverse domains and treatments. For each of them, we build the augmented context $\mathbf{x}_t^{c}$ (Eq. \ref{eq:augmentedcontext}) and run the selected \gls{CB} approaches to estimate $o_t^*$. Since the estimated $o_t^*$ leads to the indication of which individual members' opinions should be trusted or not in the \gls{HMT} for the best possible decision, we expect reductions in trust calibration distances $T$. It should be noted, however, that our goal is to evaluate the trust calibration indicator, not the actual performance of \gls{HMT}s making real-time decisions. Such interactions of \gls{HMT}s and a real-time implementation of the indicator are left for future studies.%, as discussed in Section \ref{sec:discussion}.

\section{Evaluations}
\label{sec:evaluations}

% As described in the previous section, the trust calibration indicator estimation of $o^*$ can be implemented by any \gls{CB} algorithm. The choice might consider domain knowledge, available context features and their relationships with the actions/opinions, and computational costs. However, when evaluating possible \gls{CB} algorithms, it is also important to keep in mind that, due to their dynamics, the models will change with each new data presented to them. Different from traditional supervised \gls{ML} approaches, \gls{CB} algorithms will dynamically learn the best possible opinions with respect to the context considering all previous experiences with the perceived rewards, as opposed to \gls{ML} models trained with a fixed dataset. Since no ground truth is given at each trial, only rewards, \gls{CB} algorithms will seek to maximize rewards (or minimize regrets with respect to the optimal opinions), updating their internal models at each trial $t$. Hence, the order of the trials might affect the learning dynamics and the total rewards obtained by the algorithm. Therefore, in the following evaluations, we run the \gls{CB} algorithms 100 times each, shuffling the data and reporting the sample mean and standard error of the total obtained rewards perceived by the decisions led by the opinions $o_1$, $o_2$, $o$ and estimated $o^*$.

Due to their dynamics, \gls{CB} models will change with each new data presented to them. Different from traditional supervised \gls{ML} approaches, \gls{CB} algorithms will dynamically learn the best possible opinions with respect to the context considering all previous experiences with the perceived rewards, as opposed to \gls{ML} models trained with a fixed dataset. Since no ground truth is given at each trial, only rewards, \gls{CB} algorithms will seek to maximize rewards (or minimize regrets with respect to the optimal opinions), updating their internal models at each trial $t$. Hence, the order of the trials might affect the learning dynamics and the total rewards obtained by the algorithm. Therefore, in the following evaluations, we run the \gls{CB} algorithms 100 times each, shuffling the data and reporting the sample mean and standard error of the total obtained rewards perceived with the decisions led by the opinions $o_1$, $o_2$, $o$ and estimated $o^*$.

% We start by evaluating the trust calibration indicator using the data provided by \citet{trust-lu2021}. As described before, the authors work with speed dating data, having the human subjects, assisted by \gls{AI} (implemented by \gls{ML} classifiers), predict if a person in the database would meet their date again in the future. Experimental treatments consisted in varying levels of agreement between humans and \gls{AI} and the \gls{ML} classifiers accuracy. Also, the human subjects received a monetary performance compensation bonus. As such, for each correct assessment of the future date possibility, the \gls{HMT} perceived a unit reward (it added 1 point to the total, which was later converted to money). 

We start by evaluating the trust calibration indicator using the data provided by \citet{trust-lu2021}. As described before, the authors work with speed dating data, having the human subjects, assisted by \gls{AI}, predict if a person in the database would meet their date again in the future. For each correct assessment of the future date possibility, the \gls{HMT} perceived a unit reward.% (it added 1 point to the total, which was later converted to money). 

\begin{table}
\centering
\begin{tabular}{lllll}
\toprule
\multicolumn{1}{c}{\textbf{}} & \multicolumn{3}{c}{\textbf{Level of agreement}} \\
  & \textbf{High} & \textbf{Medium} & \textbf{Low} \\
    \midrule
 Trials [$n$] & 2,400 & 3,390 & 3,240\\
 Maximum [$G(\tau)$]  & 2,400 & 3,390 & 3,240\\
 AI $o_2$ [$g(\tau)$] & 1,680 & 2,373 & 2,268 \\
 HMT $o$ [$g(\tau)$] & 1,537 & 2,175 & 2,054\\
 \hline
\multicolumn{1}{l}{\textbf{Indicator results}} & (estimated & $o^*$ [$g(\tau)$]) & \multicolumn{1}{c}{} \\
 % ~~~ (estimated $o^*$) & & & \\
 % ~~~$o^*$ mean/se) & & & \\
 CB LinUCB & 2,263.48/ & 3,270.09/ & 3,097.06/ \\
  & ~~~10.28 & ~~~79.94 & ~~~11.61 \\ 
 CB DT & 960.00/ & 1,356.00/ & 1,296.00/ \\
  & ~~~0 & ~~~0 & ~~~0 \\
 CB ANN & 1,880.20/ & 2,749.93/ & 2,549.50/ \\
  & ~~~43.62 & ~~~63.79 & ~~~62.67 \\
 \bottomrule
\end{tabular}
\caption{Trust calibration indicator results (mean/standard error of total rewards) using data from \citet{trust-lu2021}, experiment 1. CB LinUCB achieves significantly higher results than the HMT (t-tests).}
\label{tab:lu2021exp1}
\end{table}

% Table \ref{tab:lu2021exp1} shows results from the first experiments of \citet{trust-lu2021} (upper part) and the trust calibration indicator (lower part) using the same data, with estimations of $o^*$ given by \gls{LinUCB}, \gls{DT} and \gls{ANN} contextual bandits implementations. Interestingly, across the three experimental treatments (levels of agreement between humans and \gls{AI}), the total rewards from decisions led by the \gls{HMT}'s consensual opinions $o$ are slightly less than the total rewards from decisions led by the \gls{AI}'s opinions $o_2$, which indicates that the opinions of the \gls{AI} were more accurate than the consensus achieved by the \gls{HMT}. Regarding the trust calibration distances for the \gls{HMT}, calculated as Equation \ref{eq:trustcalibration}, $T = 863; 1{,}215; 1{,}186$ for high, medium and low levels of agreement, respectively. However, by estimating $o^*$ with \gls{CB} \gls{LinUCB}, the trust calibration distances are greatly reduced to $T \approx 137; 120; 143$, showing the effectiveness of the trust calibration indicator. Since the results using data from the other experiments are very similar, we display them in the Appendix.

Table \ref{tab:lu2021exp1} shows results from the first experiments of \citet{trust-lu2021} (upper part) and the trust calibration indicator (lower part) using the same data, with estimations of $o^*$ given by \gls{LinUCB}, \gls{DT} and \gls{ANN} \gls{CB} implementations. The trust calibration distances (Eq. \ref{eq:trustcalibration}) for the \gls{HMT} are $T = 863; 1{,}215; 1{,}186$ for high, medium and low levels of agreement, respectively. However, by estimating $o^*$ with \gls{CB} \gls{LinUCB}, the trust calibration distances are greatly reduced to $T \approx 137; 120; 143$, showing the effectiveness of trust calibration. Results using data from the other experiments from \citet{trust-lu2021} are very similar (see Appendix).

% The next dataset brings an example of the trust calibration indicator applied to multi-valued decision-making (as opposed to binary). As detailed in Section \ref{sec:methods:data}, \citet{teams-noti2023} asked their human subjects for likelihood values, ranging from 0 to 100 in increments of 10. In this case, the perceived reward of a decision associated with the opinion $o_t$ is given by $r(o_t, \mathbf{x}_t) = 100 - (\vert y_t - \hat{y}_t \vert)$, where, for a trial $t$, $\hat{y}_t$ is the opinion of the risk (\% likelihood of violation of release terms) issued by the agent/team and $y_t$ is the corresponding opinion leading to the ground truth (either 0\% or 100\%). For this dataset, the augmented context $\mathbf{x}_t^{(c)}$ for the trust calibration indicator at trial $t$ is composed of the classification variables $\mathbf{x}_t$ presented to the human subjects, the \gls{AI} opinion $o_{2,t}$ and the consensual decision $o_t$ of the \gls{HMT}. However, the output of the trust calibration indicator is not only which opinions should be trusted or not, but also an alternative opinion (estimated $o_t^*$) in case both $o_t$ and $o_{2,t}$ are tagged as untrustworthy for a particular trial $t$. 

The next dataset brings an example of the trust calibration indicator applied to multi-valued decision-making (as opposed to binary). In this case, the perceived reward of a decision associated with the opinion $o_t$ is given by $r(o_t, \mathbf{x}_t) = 100 - (\vert y_t - \hat{y}_t \vert)$, where, for a trial $t$, $\hat{y}_t$ is the opinion of the risk (\% likelihood of violation of release terms) issued by the agent/team and $y_t$ is the corresponding opinion leading to the ground truth (either 0\% or 100\%). For this dataset, the augmented context $\mathbf{x}_t^{(c)}$ for the trust calibration indicator at trial $t$ is composed of the classification variables $\mathbf{x}_t$ presented to the human subjects, the \gls{AI} opinion $o_{2,t}$ and the consensual decision $o_t$ of the \gls{HMT}. However, the output of the trust calibration indicator is not only which opinions should be trusted or not, but also an alternative opinion (estimated $o_t^*$) in case $o_t$, $o_{1,t}$ and $o_{2,t}$ are tagged as untrustworthy for a particular trial $t$. 

% Results for the experiments considering data from \citet{teams-noti2023} are shown in Table \ref{tab:noti2023}. This table brings the total perceived rewards, as well as \gls{HMT} rewards, separated for each treatment described in Section \ref{sec:methods:data}. However, in this case, because the perceived reward is given by $r(o_t, \mathbf{x}_t) = 100 - (\vert y_t - \hat{y}_t \vert)$, the maximum rewards are given by $G(\tau) = 100n$ in all treatments, since $R(o_t^*, \mathbf{x}_t) = 100$ when $\hat{y}_t = y_t$. As seen in the upper part of Table \ref{tab:noti2023}, the total perceived rewards of decisions led by \gls{HMT}'s $o$ are slightly lower than the total perceived rewards from decisions led by \gls{AI}'s $o_2$, showing that the \gls{AI} agent is more accurate on average than the \gls{HMT}. The trust calibration distances for the \gls{HMT} are $T = 464{,}660; 171{,}340; 180{,}030$ for the update, learned and omniscient treatments, respectively. In contrast, the estimation of $o^*$ for the trust calibration indicator results in $T \approx 355{,}300; 139{,}300; 141{,}200$ using the \gls{DT} \gls{CB} algorithm (lower part of the table). %This result can be compared with the other \gls{AI} treatments of Table \ref{tab:noti2023} by the rate of the total perceived rewards to the total possible, given by $g(\tau)/G(\tau)$. Hence, $g(\tau)/G(\tau) = 0.58; 0.59; 0.57$ for the update, learned and omniscient treatments, respectively. The trust calibration indicator leads to the higher rate of $g(\tau)/G(\tau) = 0.68$.

Results for the experiments considering data from \citet{teams-noti2023} are shown in Table \ref{tab:noti2023}. In this case, because the perceived reward is given by $r(o_t, \mathbf{x}_t) = 100 - (\vert y_t - \hat{y}_t \vert)$, the maximum rewards are given by $G(\tau) = 100n$ in all treatments, since $R(o_t^*, \mathbf{x}_t) = 100$ when $\hat{y}_t = y_t$. The trust calibration distances for the \gls{HMT} are $T = 464{,}660; 171{,}340; 180{,}030$ for the update, learned and omniscient treatments, respectively. In contrast, the estimation of $o^*$ for the trust calibration indicator results in $T \approx 355{,}300; 139{,}300; 141{,}200$ using the \gls{DT} \gls{CB} algorithm.% (lower part of the table). %This result can be compared with the other \gls{AI} treatments of Table \ref{tab:noti2023} by the rate of the total perceived rewards to the total possible, given by $g(\tau)/G(\tau)$. Hence, $g(\tau)/G(\tau) = 0.58; 0.59; 0.57$ for the update, learned and omniscient treatments, respectively. The trust calibration indicator leads to the higher rate of $g(\tau)/G(\tau) = 0.68$.

% For evaluation on this dataset, the trust calibration indicator is calculated for each decision of the update treatment because that is the traditional setup of human decision-making assisted by \gls{AI} (the other treatments assume special conditions on the \gls{AI} recommender system, as described in Section \ref{sec:methods:data}). As before, we report the mean and standard error of 100 simulations of shuffled data for each algorithm.

\begin{table}
    \centering
    \begin{tabular}{llll}
    \toprule
    \multicolumn{1}{c}{\textbf{}} & \multicolumn{3}{c}{\textbf{Treatment}} \\
    & \textbf{Update} & \textbf{Learned} & \textbf{Omniscient} \\
    \midrule     
     $n$ & 11,000 & 4,031 & 3,940 \\
     $G(\tau)$ & 1,100,000 & 403,100 & 394,000 \\
     $o_2$ [$g(\tau)$] & 663,560 & 238,970 & 235,350 \\
     $o$ [$g(\tau)$] & 635,340 & 231,760 & 213,970 \\
     \hline
     \multicolumn{1}{l}{\textbf{Indicator}} & (estimated & $o^* (g(\tau))$) & \multicolumn{1}{c}{} \\  
     % ~~~ (estimated & & & \\
     % ~~~ $o^* (g(\tau))$) & & & \\
     % ~~~$o^*$ mean/se) & & & \\
     CB LinUCB & 726,312.70/ & 256,432.70/ & 248,506.30/ \\
      & ~~~1,878.04 & ~~~705.01 & ~~~655.18\\
     CB DT & 744,700.00/ & 263,800.00/ & 252,800.00/ \\
      & ~~~0 & ~~~0 & ~~~0\\
     CB ANN & 546,799.60/ & 200,619.10/ & 217,289.10/ \\
      & ~~~11,114.83 & ~~~3,295.79 & ~~~2,767.00 \\
     \bottomrule
    \end{tabular}
    \caption{Trust calibration indicator results (mean/standard error of total rewards) using data from \citet{teams-noti2023}. CB DT achieves significantly higher results than the HMT (t-tests).}
    \label{tab:noti2023}

\end{table}

\begin{table}
    
    % \begin{tabular}{l>{\hangindent=.2cm}m{6cm}lc}
    \centering
    \begin{tabular}{ll}
    \toprule
     Number of trials [$n$] & 8,619 \\
     Maximum rewards [$G(\tau)$] & 8,619 \\
     Human $o_1$ rewards [$g(\tau)$] & 5,229 \\
     AI $o_2$ rewards [$g(\tau)$] &  6,303 \\
     HMT $o$ rewards [$g(\tau)$] & 5,979 \\
     \hline
     \textbf{Indicator results} & (estimated $o^*$ [$g(\tau)$])\\
        % ~~~(estimated $o^*$ rewards) \\
     CB LinUCB (mean/se) & 6,886.64/28.61 \\
     CB DT (mean/se) & 4,380.38/447.62 \\
     CB ANN (mean/se) & 7,857.62/16.48 \\
     \bottomrule
    \end{tabular}
    \caption{Trust calibration indicator results (mean/standard error of total rewards) using data from \citet{trust-reverberi2022}. CB ANN achieves significantly higher results than the HMT (t-tests).}
    \label{tab:reverberi2022}
\end{table}

% From Table \ref{tab:reverberi2022}, it is clear that the \gls{AI} performance would be superior without the human input. In this case, humans hinder \gls{AI} potential rewards, since the total rewards perceived by the \gls{HMT} is lower than the total rewards perceived by \gls{AI}. The trust calibration distance is given by $T = 3{,}390; 2{,}316; 2{,}640$ for the decisions led by $o_1$, $o_2$ and $o$ opinions, respectively. However, complementary performance is achieved by improved trust calibration. As seen in the lower part of Table \ref{tab:reverberi2022}, the trust calibration indicator reduces the trust calibration distance to $T \approx 761$ using \gls{CB} \gls{ANN}.

Finally, we evaluate the trust calibration indicator using data from \citet{trust-reverberi2022}, which contains diagnoses of colorectal lesions given by endoscopists assisted by \gls{AI}, therefore \gls{HMT}s. A correct diagnosis was rewarded by 1 point and an incorrect was penalized by 1 point. Results are summarized in Table \ref{tab:reverberi2022}. The trust calibration distance is given by $T = 3{,}390; 2{,}316; 2{,}640$ for the decisions led by $o_1$, $o_2$ and $o$ opinions, respectively. However, complementary performance is achieved by improved trust calibration (lower part of Table \ref{tab:reverberi2022}), with the trust calibration distance reduced to $T \approx 761$ using \gls{CB} \gls{ANN}.

As shown, the trust calibration indicator successfully reduces the trust calibration distance, leading to complementary performance. Overall, the total rewards increased by 10 to 38\%. For each dataset, one of the \gls{CB} algorithms outperformed the others in estimating $o^*$: \gls{CB} \gls{ANN} excelled with data from \citet{trust-reverberi2022}; \gls{CB} \gls{DT} was superior with data from \citet{teams-noti2023}; and \gls{CB} \gls{LinUCB} was the best approach with data from \citet{trust-lu2021}. This is expected since the domains, variables and decisions are different in each dataset.

\section{Discussion and conclusion}
\label{sec:discussion}

The key insight of our trust calibration framework is the distinction between opinions and decisions. We notice that before issuing a decision, each opinion is judged with respect to the current decision variables in order to derive a global team opinion that will subsequently translate into a decision. This internal process of trust calibration is highly subjective due to the nature of individual opinion-formation processes of each member. Despite its subjective nature, we model the trust calibration process externally by anticipating which opinions can be trusted to achieve the best outcome.% on a given decision. Moreover, trust is dynamically calibrated at each decision, with respect to opinions given for the variables of that specific decision, independently of inherent characteristics of members.

% In addition to demonstrating the critical role of trust calibration in achieving complementary performance, our framework has broader implications. First, it shifts focus away from the subjective traits of individual team members, centering solely on their capacity for contributions. Consequently, even members with lower performance have their opinions considered when building consensus. Second, our framework considers trust calibration as a bidirectional process: it refers to both human trust in \gls{AI} as well as \gls{AI} trust in humans. This dual perspective allows for the examination of human decision-making supported by \gls{AI} and enables \gls{AI} to make informed decisions based on human input. Ultimately, the trust calibration indicator identifies situations in which, for the optimal outcome, human opinions should be overridden to avoid human error.

% It is important to emphasize that

The proposed trust calibration indicator is oblivious to the models governing the formation of individual \gls{HMT} agents' opinions, including the \gls{AI}'s implementation. As such, it can be applied to any \gls{AI}-supported decision-making scenario, irrespective of the \gls{AI} technology. However, one limitation of our approach is that the indicator relies on the estimation of $o^*$ by a \gls{CB} algorithm. Our findings indicate that no single algorithm consistently outperforms others across the examined domains. This variability suggests that the trust calibration indicator estimation of $o^*$ by \gls{CB} is domain-dependent and influenced by the corresponding rewards structure. As with any \gls{ML} deployment, practical implementations of the indicator should investigate the most appropriate \gls{CB} algorithm for estimating $o^*$ on a given domain. Therefore, prior datasets or simulated data should be utilized to evaluate models before deployment, also mitigating the cold start problem (limited prior information). %Another limitation of our framework is the assumption that the \gls{CB} algorithms are always able to perceive the rewards of decisions associated with its estimated opinion $o^*$. This might not be the case when there is no immediate feedback from the decision environment.

A core assumption of our framework is that reward feedback for each decision is immediately and accurately observable, enabling timely updates of the trust calibration indicator via \gls{CB}. While this assumption holds in controlled experimental environments or domains with clear, instantaneous outcomes, it may not apply universally, since reward signals can be significantly delayed, noisy, or unavailable. Addressing these challenges requires adaptations of our method, such as integrating delayed calibration, or leveraging proxy or intermediate feedback signals. Those are important directions for future work to broaden the framework’s practical applicability.

Finally, our trust calibration framework opens several avenues for future research and application. For example, it can be applied to team composition analysis to evaluate individual contributions or integrated into real-time decision support to guide trust behavior by displaying the indicator prior to decisions. However, realizing these applications will require future studies involving human participants to identify effective ways to present the indicator to \gls{HMT}s. This next step is essential to gain deeper insights into the practical usability, user acceptance, and overall effectiveness of our framework in live operational settings.

%%%%%%%%%%%%%%%%%%%%%%%%%%%%%%%%%%%%%%%%%%%%%%%%%%%%%%%%%%%%%%%%%%%%%%%%

%%% Use this environment to include acknowledgements (optional).
%%% This will be omitted in doubleblind mode.

% \begin{ack}
% By using the \texttt{ack} environment to insert your (optional) 
% acknowledgements, you can ensure that the text is suppressed whenever 
% you use the \texttt{doubleblind} option. In the final version, 
% acknowledgements may be included on the extra page intended for references.
% \end{ack}

%%%%%%%%%%%%%%%%%%%%%%%%%%%%%%%%%%%%%%%%%%%%%%%%%%%%%%%%%%%%%%%%%%%%%%%%

%%% Use this command to include your bibliography file.
\clearpage
\bibliography{references}
\clearpage

\section{Appendix}
\label{sec:appendix}

In this Appendix, we present the trust calibration indicator evaluations using the dataset from \citet{trust-lu2021}, experiments 2 (Tables \ref{tab:lu2021exp2-a} and \ref{tab:lu2021exp2-b}) and 3 (Tables \ref{tab:lu2021exp3-a} and \ref{tab:lu2021exp3-b}). For the evaluations using experiments 1 data, please refer to the Evaluations section of the main paper.

The second experiment conducted by \citet{trust-lu2021} had a $2 \times 2$ factorial design, varying levels of agreement of humans and the underlying \gls{ML} classifiers accuracy. Results are shown in the upper parts of Tables \ref{tab:lu2021exp2-a} and \ref{tab:lu2021exp2-b}. It is remarkable, in these cases, that the \gls{HMT} consensual opinions $o$ led to the perception of more rewards than the decisions led by the \gls{AI} opinions $o_2$ in all four treatments. However, performance would be greatly increased by improved trust calibration, as shown by the rewards from decisions led by $o^*$ (means and standard errors in the lower parts of Tables \ref{tab:lu2021exp2-a} and \ref{tab:lu2021exp2-b}). In these cases, the trust calibration distances for the \gls{HMT} are $T = 1{,}168; 1{,}407; 1{,}250; 1{,}516$ for the agreement/accuracy treatments of 80/80, 80/50, 50/80 and 50/50, respectively. In contrast, trust calibration distance is reduced to $T \approx 166; 188; 126; 156$ with the estimations of $o^*$ given by \gls{CB} \gls{LinUCB} in the trust calibration indicator.

\begin{table}

\centering
\begin{tabular}{llll}
\toprule
\multicolumn{1}{c}{\textbf{}} & \multicolumn{2}{c}{\textbf{Agreement \%/Accuracy \%}} \\
  & \textbf{80/80} & \textbf{80/50}\\
    \midrule
Trials [$n$] & 3,030 & 3,630\\
Maximum [$G(\tau)$] & 3,030 & 3,630\\
 AI $o_2$ [$g(\tau)$] & 1,717 & 2,057\\
 HMT $o$ [$g(\tau)$] & 1,862 & 2,223\\
 \hline
 % \multicolumn{1}{l}{\textbf{Indicator results}} & \multicolumn{3}{c}{} \\
 % ~~~ (estimated $o^*$ rewards) & & & \\
 \multicolumn{1}{l}{\textbf{Indicator results}} & (estimated & $o^*$ [$g(\tau)$]) & \multicolumn{1}{c}{} \\
 CB LinUCB & 2,864.34/12.39 & 3,442.53/14.54\\
 CB DT & 1,212.00/0 & 1,452.00/0\\
 CB ANN & 2,422.61/56.11 & 2,727.35/77.04\\ 
\bottomrule
\end{tabular}
\caption{Trust calibration indicator results (mean/standard error) using data from \citet{trust-lu2021}, experiment 2, Agreement \% / Accuracy \% treatments of 80/80 and 80/50. CB LinUCB achieves significantly higher results than the HMT (t-tests).}
\label{tab:lu2021exp2-a}
\end{table}

\begin{table}

\centering
\begin{tabular}{llll}
\toprule
\multicolumn{1}{c}{\textbf{}} & \multicolumn{2}{c}{\textbf{Agreement \%/Accuracy \%}} \\
  & \textbf{50/80} & \textbf{50/50}\\
    \midrule
 Trials [$n$] & 3,390 & 3,930 \\
 Maximum [$G(\tau)$] & 3,390 & 3,930\\
 AI $o_2$ [$g(\tau)$] & 1,921 & 2,227 \\
 HMT $o$ [$g(\tau)$] & 2,140 & 2,414\\
 \hline
 % \multicolumn{1}{l}{\textbf{Indicator results}} & \multicolumn{3}{c}{} \\
 % ~~~ (estimated $o^*$ rewards) & & & \\
 \multicolumn{1}{l}{\textbf{Indicator results}} & (estimated & $o^*$ [$g(\tau)$]) & \multicolumn{1}{c}{} \\
 CB LinUCB & 3,263.83/90.77 & 3,774.43/12.23\\
 CB DT & 1,356.00/0 & 1,572.00/0\\
 CB ANN & 2,561.46/71.20 & 3,051.70/82.55\\ 
\bottomrule
\end{tabular}
\caption{Trust calibration indicator results (mean/standard error) using data from \citet{trust-lu2021}, experiment 2, Agreement \% / Accuracy \% treatments of 50/80 and 50/50. CB LinUCB achieves significantly higher results than the HMT (t-tests).}
\label{tab:lu2021exp2-b}
\end{table}

The final experiment of \citet{trust-lu2021} varies human confidence in their decisions when they agree and when they disagree with the \gls{AI} opinions. Tables \ref{tab:lu2021exp3-a} and \ref{tab:lu2021exp3-b} show the results of the trust calibration indicator across the four treatments considered by the authors. As shown in the upper parts of the tables, the total rewards perceived by decisions led by the \gls{HMT} consensual $o$ are less than the total rewards perceived by the decisions led by \gls{AI}'s $o_2$ in all cases. Regarding the trust calibration distance for opinions $o$, $T = 743; 624; 752; 777$ for the treatments of high confidence agreement, low confidence agreement, high confidence disagreement and low confidence disagreement, respectively. Again, the trust calibration indicator, using $o^*$ estimated by \gls{CB} \gls{LinUCB}, reduces the trust calibration distances to $T \approx 35; 11; 18; 10$.

\begin{table}
  
  \centering
  % \begin{tabular}{l>{\hangindent=.2cm}m{6cm}lc}
\begin{tabular}{llll}
  \toprule
  \multicolumn{1}{c}{\textbf{}} & \multicolumn{2}{c}{\textbf{Treatment}} \\
    & \textbf{High confidence} & \textbf{Low confidence } \\
    & \textbf{agreement} & \textbf{agreement} \\
   \midrule
   Trials [$n$]  & 2,288 & 1,958\\
   Maximum [$G(\tau)$] & 2,288 & 1,958\\
   AI $o_2$ [$g(\tau)$] & 1,768 & 1,513\\
   HMT $o$ [$g(\tau)$] & 1,545 & 1,334\\
   \hline
 %   \multicolumn{1}{l}{\textbf{Indicator results}} & \multicolumn{3}{c}{} \\
 % ~~~ (estimated $o^*$ rewards) & & & \\ 
 \multicolumn{1}{l}{\textbf{Indicator results}} & (estimated & $o^*$ [$g(\tau)$]) & \multicolumn{1}{c}{} \\
   CB LinUCB & 2,253.13/4.80 & 1,947.10/0.22\\
   CB DT & 832.00/0 & 712.00/0\\
   CB ANN & 1,910.67/40.11 & 1,643.68/34.71\\
   
  \bottomrule
  \end{tabular}
  \caption{Trust calibration indicator results (mean/standard error) using data from \citet{trust-lu2021}, experiment 3, agreements treatments. CB LinUCB achieves significantly higher results than the HMT (t-tests).}
  \label{tab:lu2021exp3-a}
  \end{table}

\begin{table}
  
  \centering
  % \begin{tabular}{l>{\hangindent=.2cm}m{6cm}lc}
\begin{tabular}{llll}
  \toprule
  \multicolumn{1}{c}{\textbf{}} & \multicolumn{2}{c}{\textbf{Treatment}} \\
    & \textbf{High confidence} & \textbf{Low confidence} \\
    & \textbf{disagreement} & \textbf{disagreement} \\
   \midrule
   Trials [$n$] & 2,244 & 2,354\\
   Maximum [$G(\tau)$] & 2,244 & 2,354\\
   AI $o_2$ [$g(\tau)$] & 1,734 & 1,819 \\
   HMT $o$ [$g(\tau)$] & 1,492 & 1,577\\
   \hline
 %   \multicolumn{1}{l}{\textbf{Indicator results}} & \multicolumn{3}{c}{} \\
 % ~~~ (estimated $o^*$ rewards) & & & \\
 \multicolumn{1}{l}{\textbf{Indicator results}} & (estimated & $o^*$ [$g(\tau)$]) & \multicolumn{1}{c}{} \\
   CB LinUCB  & 2,225.98/2.28 & 2,344.45/0.24\\
   CB DT  & 816.00/0 & 856.00/0 \\
   CB ANN & 1,780.35/51.08 & 1,905.66/48.43\\
   
  \bottomrule
  \end{tabular}
  \caption{Trust calibration indicator results (mean/standard error) using data from \citet{trust-lu2021}, experiment 3, disagreement treatments. CB LinUCB achieves significantly higher results than the HMT (t-tests).}
  \label{tab:lu2021exp3-b}
  \end{table}

\end{document}